%% file: paper.tex
\documentclass[11pt]{article}

\usepackage[preprint]{acl} 
\usepackage{times}
\usepackage{latexsym}
\usepackage[utf8]{inputenc} 
\usepackage[T1]{fontenc}    
\usepackage{booktabs}       
\usepackage{amsfonts}       
\usepackage{amsmath}
\usepackage{nicefrac}       
\usepackage{microtype}      
\usepackage{inconsolata}
\usepackage{graphicx}
\usepackage{subcaption}
\usepackage{pifont}
\usepackage{multirow} 
\usepackage{enumitem}
\newcommand{\cmark}{\ding{51}}
\newcommand{\xmark}{\ding{55}}

\usepackage[nameinlink]{cleveref}
\title{Do Clinical Models Change Treatment Decisions?}

\author{
Dongkyu Cho$^{1}$, Miao Zhang$^{1}$, Rumi Chunara$^{1}$ \\
$^{1}$ New York University \\
\textit{dongkyu.cho@nyu.edu} \quad \textit{miaozhng@nyu.edu},  \quad \textit{rumi.chunara@nyu.edu}
}

\begin{document}

\maketitle

\begin{abstract}
Clinical foundation models are evaluated with factual or exam-style
medical QA, but treatment decisions must change when patient context changes.
We introduce ClinPivot, an auditable treatment-decision benchmark built from
biomedical relations and pivoted patient contexts. ClinPivot asks whether
models change treatment choices when new clinical constraints shift the action
space. We find that strong medical QA performance does not reliably predict
decision-making performance: frontier models and task-adapted Qwen variants
often fail to change decisions correctly, and model rankings shift across
evaluation regimes. Decision-structured supervision improves pivot-sensitive
decision-making and medical QA under matched knowledge budgets, while
lightweight replay reduces losses in general assistant ability.
\end{abstract}

\begin{figure}[!t]
\centering
\includegraphics[width=\linewidth]{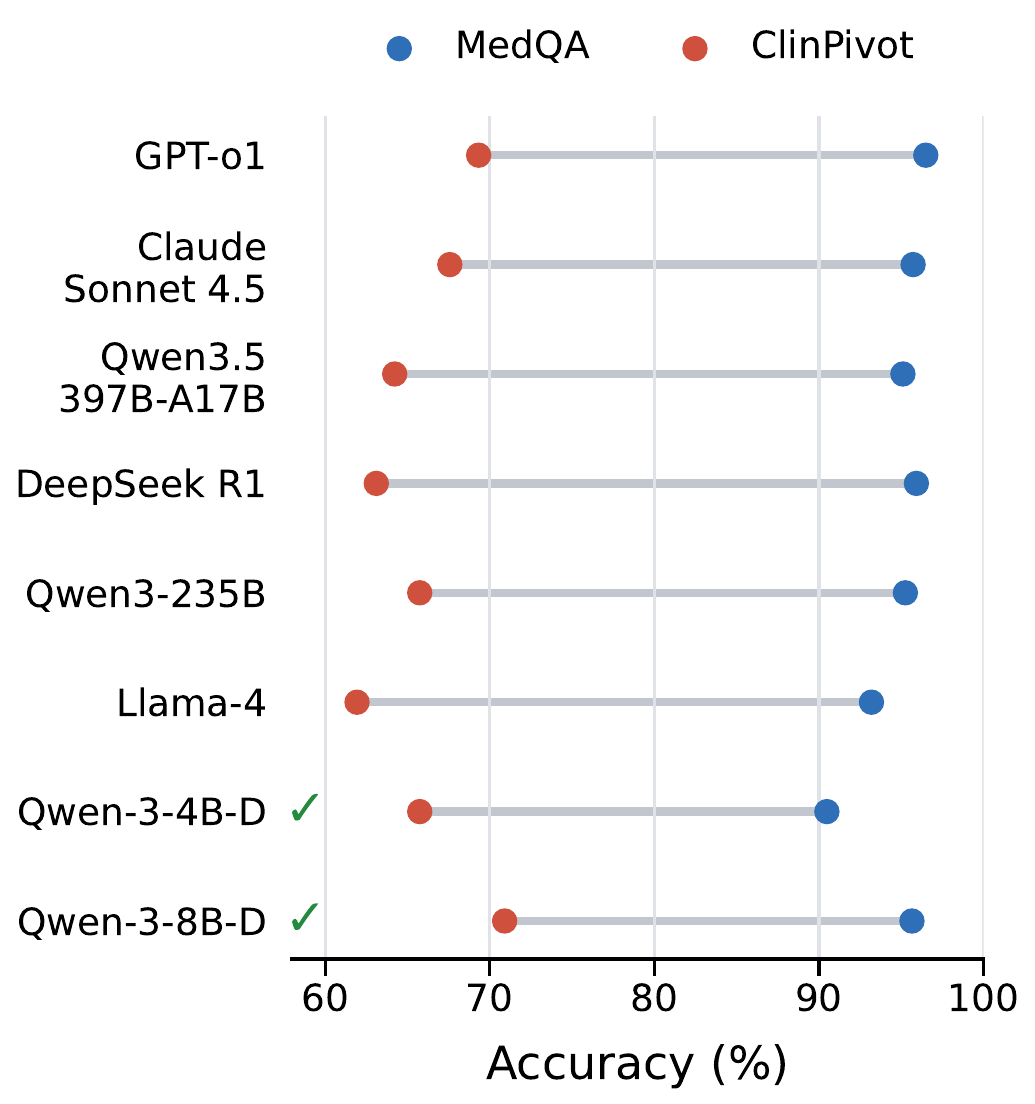}
\caption{\textbf{Models with high MedQA accuracy still show large drops on
\textsc{ClinPivot}.} Unmarked frontier-model accuracies are from
\Cref{tab:frontier_results}; green check marks indicate decision-trained Qwen
models from \Cref{tab:fine_tuning_results}. Values are accuracies in percent.}
\label{fig:frontier_results}
\end{figure}

\section{Introduction}
\input{sections/introduction}

\section{Related Work}
\input{sections/related_works}

\section{\textsc{ClinPivot}: Treatment Decisions Under Constraints}
\input{sections/method}

\section{Experiments}
\subsection{Experimental Setting}
\input{sections/setting}
\subsection{Experimental Results}
\input{sections/results}

\section{Concluding Remarks}
\input{sections/conclusion}

\clearpage

\section*{Limitations}
\input{sections/limitations}

\section*{Ethical considerations}
\input{sections/ethical_considerations}

\bibliography{main}

\appendix

\input{sections/appendix}


\end{document}

%% file: sections/introduction.tex
Medical QA benchmarks such as MedQA~\citep{jin2020diseasedoespatienthave}
and PubMedQA~\citep{jin-etal-2019-pubmedqa} have become a standard way to
evaluate clinical language models. They are useful, but they mostly reward
models for answering fixed questions, where the relevant facts and answer set
do not change. Treatment decisions have a different failure mode: a drug can be
appropriate for a disease and still become the wrong choice once the patient
has a contraindicating comorbidity, an interacting medication, a relevant
allergy, or another constraint. Recent work likewise cautions that strong
medical QA performance does not settle whether models reason reliably in
clinical settings~\citep{singhal2023large,tu2023generalistbiomedicalai,griot2025large,kim2025limitations}.

The gap is not simply knowledge, but conditional action. A model may know that
drug $A$ treats disease $D$, while still choosing $A$ in a case where $A$ has
become unsafe or inappropriate. We ask a direct question: do clinical models
change treatment decisions when the patient context changes? This framing
follows clinical reasoning work that emphasizes decisions under partial
information and patient-specific constraints~\citep{ten2017principles,wu2024clinical}.
\Cref{fig:frontier_results} previews the central pattern: models with high
medical QA scores still drop sharply on pivoted treatment decisions.

We introduce \textsc{ClinPivot}, an auditable clinical treatment-decision
benchmark built from PrimeKG-style disease, drug, phenotype, indication,
contraindication, interaction, and off-label relations~\citep{primekg}. Each
example presents a short patient vignette and candidate treatments. The
vignette contains a decision pivot: a clinically meaningful change in patient
state that can alter or constrain the treatment choice. The model must select
the appropriate action in the pivoted case, rather than merely recall a
disease--drug association.

\textsc{ClinPivot} keeps this behavior inspectable. Labels come from graph
operations rather than another language model, and each item stores the disease
node, candidate drug nodes, pivot type, banned drugs when applicable, and
supporting evidence triples. This structure lets us measure both decision
accuracy and pivot sensitivity: whether a model avoids the treatment made
unsafe or inappropriate by the new constraint.

Our experiments show that high medical QA performance does not reliably imply
pivot-sensitive decision-making. Strong frontier models and task-adapted Qwen
variants can do well on MedQA while failing to update treatment choices under
decision pivots, and model rankings change across evaluation regimes. We also
find that decision-structured supervision improves \textsc{ClinPivot} more
directly than QA-only supervision under matched clinical knowledge budgets.
Because this training can erode broad assistant abilities, we evaluate general
capability retention and show that lightweight replay preserves much of the
base model's general behavior while retaining most clinical decision gains.

\paragraph{Contributions.}
We make the following contributions:
\begin{itemize}[nosep,leftmargin=*]
    \item We introduce \textsc{ClinPivot}, a clinical benchmark that tests
    whether models change treatment choices when patient constraints change.
    
    \item We build the examples from biomedical graph relations and keep the
    supporting evidence for each item.
    
    \item We show that medical QA scores do not reliably predict performance
    on pivoted treatment decisions, and that model rankings change across the
    two settings.
    
    \item We compare QA-style and decision-style training, and show that simple
    replay helps preserve general abilities during clinical decision training.
\end{itemize}

%% file: sections/related_works.tex
\paragraph{Clinical reasoning models.}
Large language models have shown strong performance on medical knowledge and
clinical reasoning tasks, including Med-PaLM-style medical QA and generalist
biomedical AI systems~\citep{singhal2023large,tu2023generalistbiomedicalai}.
However, recent work argues that high benchmark accuracy does not guarantee
reliable clinical reasoning: models may lack metacognitive awareness, exhibit
inflexible reasoning, or fail in operational clinical settings even when they
answer medical questions correctly~\citep{griot2025large,kim2025limitations,jiang2025generalistfoundationmodelsclinical}.
\textsc{ClinPivot} follows this line of work but focuses on a specific,
auditable behavior: whether a model changes a treatment decision when a
patient-specific constraint changes the action space.

\paragraph{Clinical benchmarks.}
Medical benchmarks such as MedQA and PubMedQA have become standard tools for
evaluating biomedical knowledge and question
answering~\citep{jin2020diseasedoespatienthave,jin-etal-2019-pubmedqa}. These
benchmarks are useful, but they generally present fixed questions with fixed
answer choices. They do not directly test whether a model revises an action when
the patient context changes. \textsc{ClinPivot} is complementary: it uses
PrimeKG-style biomedical relations~\citep{primekg} to construct
treatment-selection examples with decision pivots, following the broader idea
that correct reasoning often depends on identifying pivotal state
changes~\citep{cho2026correctreasoningpathsvisit}. This lets us evaluate not
only whether a model knows a disease--treatment association, but whether it
treats that association as conditional on patient-specific constraints.

%% file: sections/method.tex
\paragraph{The task.}
\textsc{ClinPivot} evaluates treatment selection after the patient context
changes. Let $d$ be a disease and $a_0$ an indicated treatment for $d$.
A decision pivot $p$ is a minimal addition to the patient state, such as a
contraindicating comorbidity, interacting medication, allergy, or off-label
tension, following prior work on decision pivots in
reasoning~\citep{cho2026correctreasoningpathsvisit}. The pivot induces a new
decision state $(d,p)$ in which $a_0$ may no
longer be the appropriate graph-derived action. Given a vignette for $(d,p)$
and a candidate set $C$ containing plausible treatments, the model must select
the gold treatment under graph-derived constraints, $y \in C$. This setup
tests whether the model keeps choosing the original disease--treatment
association after the action space has changed.

\paragraph{Building examples.}
PrimeKG-style biomedical knowledge graphs provide the relations needed to build
these cases at scale~\citep{primekg}. Indication edges identify
candidate treatments for a disease. Contraindication and interaction edges
define hard constraints that can make an otherwise plausible treatment unsafe.
Off-label relations and pharmacologically nearby drugs provide tempting
alternatives. For each eligible disease, we sample an original treatment
candidate $a_0$, instantiate one pivot family, and recompute the gold treatment:
if the pivot bans $a_0$, the graph-derived gold is an indicated drug that is
not banned; if the pivot is an off-label tradeoff, the indicated treatment
remains the graph-derived gold and the off-label drug becomes a distractor.
Disease feature text and phenotype edges supply the vignette, while indicated,
contraindicated, off-label, and nearby drugs supply plausible answer choices.

\paragraph{Evidence and checks.}
Each row contains the benchmark prompt, answer choices, graph-derived gold
answer, and a reference rationale. The rationale is not treated as a human
chain of thought; it is a structured, graph-aligned explanation derived from
the same evidence used to construct the item. Metadata stores the disease node,
gold drug node, candidate drug nodes, pivot type, banned drugs, and evidence
triples. For release artifacts, we apply a reject-only LLM-assisted clinical
consistency screen after rule-based generation. The screen can remove
incoherent examples, but it cannot create labels or change the graph-derived
gold treatment.
\textsc{ClinPivot} is not intended to certify clinical safety or replace
guideline-based evaluation; it isolates one capability: updating treatment
decisions when structured patient constraints change. Additional construction
and validation details are in Appendix~\ref{app:generation}.

%% file: sections/setting.tex
\paragraph{Evaluation tasks.}
We evaluate models along three axes: standard medical QA, pivot-sensitive
clinical decision making, and general capability retention. MedQA and PubMedQA
measure exam-style or abstract-grounded biomedical QA, while \textsc{ClinPivot}
asks models to select a treatment under shifts in patient-specific clinical
constraints.

\paragraph{Models.}
We first evaluate strong general-purpose reasoning models, including GPT o1,
Claude Sonnet 4.5, Qwen3.5-397B-A17B-FP8, and DeepSeek
R1~\citep{openai2024o1systemcard,anthropic2025claudesonnet45,qwen2026qwen35modelcard,deepseekai2025deepseekr1}.
Qwen3.5-397B-A17B-FP8 is a sparse model with 397B total parameters and 17B
activated parameters.
We also evaluate two additional open-weight frontier models,
Qwen3-235B-A22B and
Llama-4-Maverick-17B-128E~\citep{yang2025qwen3,meta2025llama4maverick}, using
the same evaluation protocol. For controlled training
experiments, we use the Hugging Face checkpoints \texttt{Qwen/Qwen3-4B} and
\texttt{Qwen/Qwen3-8B}, so that we can compare training recipes while holding
the model family fixed.

\paragraph{Training and scoring.}
To test whether the decision gap can be reduced, we fine-tune Qwen-3 models
with matched clinical facts rendered either as QA examples or as
\textsc{ClinPivot}-style decision examples. We also evaluate mixed QA+Decision
training and replay variants that add general instruction and reasoning
examples, equal in size to the clinical fine-tuning data, sampled from the
substitute pretraining mixture
\texttt{allenai/dolma3\_mix-6T}~\citep{olmo2025olmo3}. All task scores are
reported as accuracy in percent. A response is correct if the selected option
matches the graph-derived gold label after normalizing option letters and
option text; free-form explanations are ignored unless they change the final
selected answer. Pivot sensitivity measures whether the model avoids the
pre-pivot treatment when that treatment is explicitly banned by metadata.
Appendix~\ref{app:experimental-details} gives the full training, retention, and
scoring protocol.

%% file: sections/results.tex
\paragraph{Medical QA performance does not imply clinical decision competence.}
\Cref{tab:frontier_results} and \Cref{fig:frontier_results} compare strong
general-purpose models on medical QA and \textsc{ClinPivot}. MedQA accuracy
ranges from 93.19 to 96.49 and PubMedQA accuracy from 75.23 to 81.98, but
\textsc{ClinPivot} scores remain much lower. The mean gaps from MedQA and
PubMedQA to \textsc{ClinPivot} are 29.97 and 14.31 points, respectively.
Across all completed \textsc{ClinPivot} runs, scores range from 61.93 to 69.32.
Rankings also shift: DeepSeek R1 has the second-highest MedQA score but the
second-lowest \textsc{ClinPivot} score among models with both metrics, while
Claude Sonnet 4.5 and Qwen3.5-397B-A17B-FP8 are lower on MedQA but higher on
\textsc{ClinPivot}.
The benchmark therefore tests whether a model updates an initially plausible
treatment decision when a patient-specific pivot changes the clinical
constraints.

\begin{table}[t]
\centering
\caption{\textbf{Medical QA scores are high, but \textsc{ClinPivot} scores remain
substantially lower across frontier models.} \textsc{ClinPivot} scores are
computed on the full 2{,}015-example test set.}
\label{tab:frontier_results}
\resizebox{\linewidth}{!}{%
\begin{tabular}{llccc}
\toprule
\textbf{Model} & \textbf{Access} & \textbf{MedQA} & \textbf{PubMedQA} & \textbf{\textsc{ClinPivot}} \\
\midrule
GPT o1       & Closed API & 96.49 & 81.98 & 69.32 \\
Claude Sonnet 4.5   & Closed API & 95.72 & 80.57 & 67.57 \\
Qwen3.5-397B-A17B-FP8 & Open-weight & 95.10 & 79.31 & 64.22 \\
DeepSeek R1  & Open-weight & 95.92 & 80.29 & 63.10 \\
\midrule
Qwen3-235B-A22B & Open-weight & 95.25 & 80.38 & 65.74 \\
Llama-4-Maverick & Open-weight & 93.19 & 75.23 & 61.93 \\
\bottomrule
\end{tabular}
}
\end{table}

\begin{figure}[t]
\centering
\includegraphics[width=\linewidth]{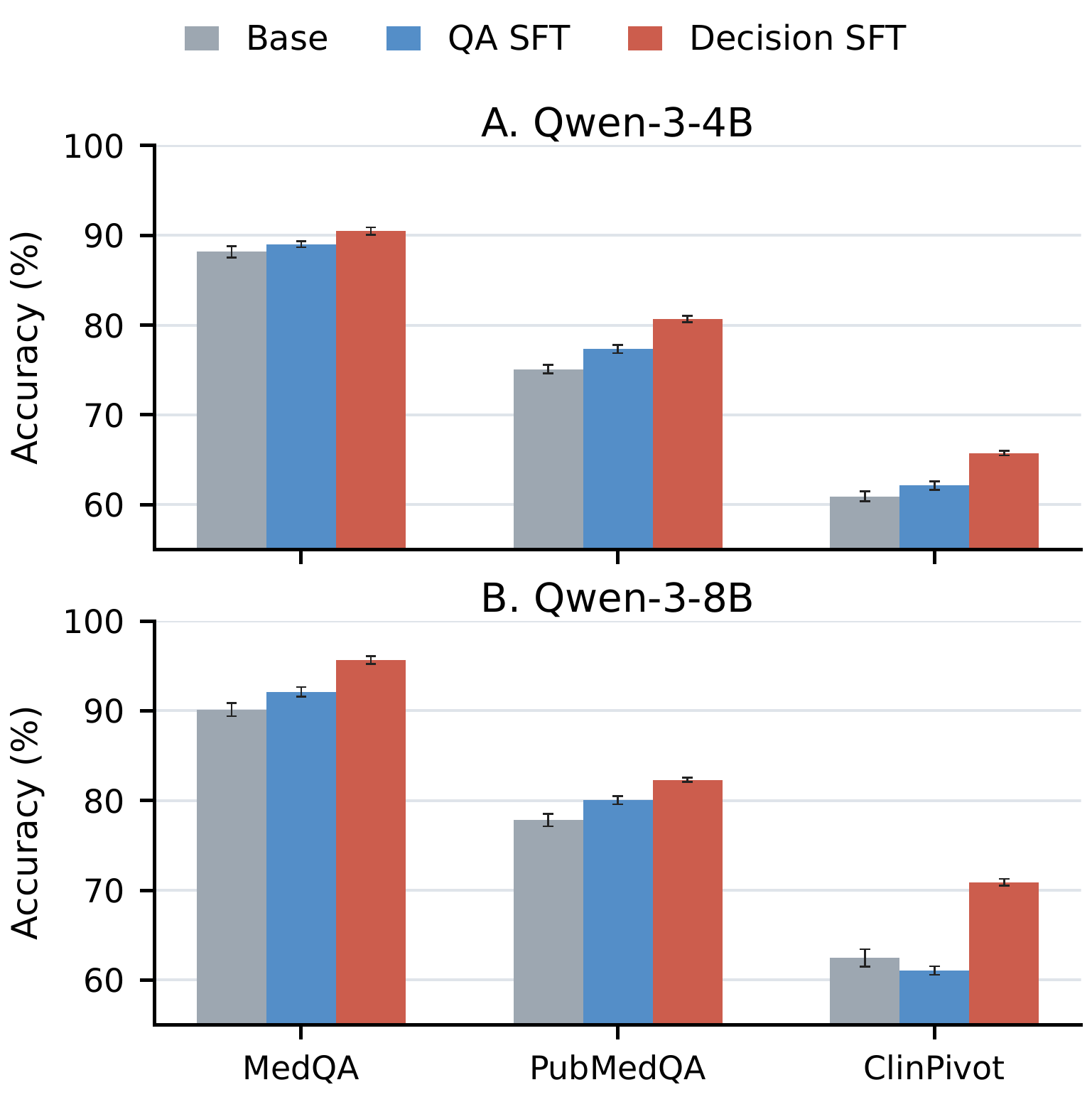}
\caption{\textbf{Decision-structured supervision improves \textsc{ClinPivot} more than
QA-only supervision.} Panels A and B report Qwen-3-4B and Qwen-3-8B results from
\Cref{tab:fine_tuning_results} across MedQA, PubMedQA, and \textsc{ClinPivot};
error bars show standard error across five random seeds.}
\label{fig:fine_tuning_results}
\end{figure}

\paragraph{Decision-structured supervision is more effective than QA-only supervision.}
\Cref{tab:fine_tuning_results} and \Cref{fig:fine_tuning_results} isolate the
effect of the training signal using Qwen-3 models. Decision SFT consistently
outperforms QA SFT on \textsc{ClinPivot}: it improves Qwen-3-4B by 4.82 points
over base, and Qwen-3-8B by 8.42 points. QA SFT is weaker and can even reduce
decision accuracy on the larger model despite improving MedQA. Thus, the format
of clinical supervision matters: decision examples train the model to connect
facts to actions under constraints, not only to recall biomedical associations.

\begin{table}[t]
\centering
\caption{\textbf{Decision-structured supervision improves treatment-decision
accuracy across both Qwen-3 model sizes.} Scores are mean $\pm$ standard error
across five random seeds.}
\label{tab:fine_tuning_results}
\resizebox{\linewidth}{!}{%
\begin{tabular}{lcccc}
\toprule
\textbf{Model} & \textbf{Fine-tuned} & \textbf{MedQA} & \textbf{PubMedQA} & \textbf{\textsc{ClinPivot}} \\
\midrule
Qwen-3-4B       & \xmark & $88.17{\pm}0.64$ & $75.10{\pm}0.48$ & $60.92{\pm}0.56$ \\
Qwen-3-4B (QA)     & \cmark & $89.02{\pm}0.33$ & $77.35{\pm}0.45$ & $62.13{\pm}0.46$ \\
Qwen-3-4B (Decision)     & \cmark & $90.48{\pm}0.43$ & $80.69{\pm}0.35$ & $65.74{\pm}0.25$ \\
\midrule
Qwen-3-8B       & \xmark & $90.13{\pm}0.73$ & $77.82{\pm}0.68$ & $62.48{\pm}0.97$ \\
Qwen-3-8B (QA)     & \cmark & $92.12{\pm}0.53$ & $80.04{\pm}0.47$ & $61.07{\pm}0.48$ \\
Qwen-3-8B (Decision)     & \cmark & $95.65{\pm}0.44$ & $82.31{\pm}0.23$ & $70.90{\pm}0.36$ \\

\bottomrule
\end{tabular}
}
\end{table}

\paragraph{Clinical decision training creates a retention tradeoff.}
\Cref{tab:continual_learning} examines the effect of clinical decision training
on general capability retention. Decision
SFT improves Qwen-3-8B by 8.42 points on decision accuracy and 20.49 points on
pivot sensitivity, while QA + Decision SFT reaches the strongest clinical
scores. These gains come with a cost: Decision SFT reduces IFEval by 7.88 points
and average retention by 6.27 points, with larger losses for QA + Decision SFT.
Clinical decision training can therefore improve pivot-sensitive reasoning
while weakening general assistant behavior. Full per-task retention scores are
reported in \Cref{tab:retention_full}, whose aggregate column matches the
Avg. Ret. column in \Cref{tab:continual_learning}.

\paragraph{Replay preserves general capabilities while retaining most clinical
gains.}
Replay mitigates this tradeoff without erasing most clinical gains. Decision
SFT + Replay reaches 70.12 decision accuracy and 60.73 pivot sensitivity, close
to Decision SFT without replay, while reducing the retention loss to 1.69 points
on IFEval and 1.23 points on average retention. This suggests that training
clinical decision models should account for both domain-specific decision
competence and general assistant reliability.

\begin{table}[t]
\centering
\small
\caption{
\textbf{Clinical decision training improves treatment-decision performance but can
degrade general capabilities; replay mitigates this tradeoff.} Avg. Ret.
summarizes ARC, TruthfulQA, WinoGrande, GSM8K, HellaSwag, and MMLU; full
per-task scores are reported in \Cref{tab:retention_full}.
}
\label{tab:continual_learning}
\resizebox{\linewidth}{!}{%
\begin{tabular}{lcccccc}
\toprule
\multirow{2}{*}{Training Data} 
& \multicolumn{2}{c}{Clinical QA} 
& \multicolumn{2}{c}{Clinical Decision} 
& \multicolumn{2}{c}{General Retention} \\
\cmidrule(lr){2-3} \cmidrule(lr){4-5} \cmidrule(lr){6-7}
& MedQA & PubMedQA 
& Dec. Acc. & Pivot Sens. 
& IFEval & Avg. Ret. \\
\midrule
Base model (Qwen-3-8B)
& 90.13 & 77.82 & 62.48 & 41.35 & 74.60 & 66.10 \\
QA SFT 
& 92.12 & 80.04 & 61.07 & 39.92 & 71.35 & 63.42 \\
Decision SFT 
& 95.65 & 82.31 & 70.90 & 61.84 & 66.72 & 59.83 \\
QA + Decision SFT 
& 95.94 & 83.10 & 72.18 & 64.27 & 65.81 & 58.96 \\
Decision SFT + Replay 
& 95.21 & 81.94 & 70.12 & 60.73 & 72.91 & 64.87 \\
QA + Decision SFT + Replay 
& 95.73 & 82.76 & 71.64 & 63.05 & 72.28 & 64.41 \\
\bottomrule
\end{tabular}
}
\end{table}

%% file: sections/conclusion.tex
\textsc{ClinPivot} turns a simple clinical expectation into a benchmark
question: when the patient context changes, does the model change the
treatment? Our results suggest that medical QA accuracy is not a reliable proxy
for this behavior. Decision-structured supervision helps, and replay preserves
much of the general capability that can be lost during clinical decision
training.

%% file: sections/limitations.tex
\textsc{ClinPivot} is a benchmark, not a test of clinical safety. Its labels
come from biomedical graph relations, so they can be inspected, but they also
inherit missing links, noisy relations, and outdated source information. The
gold treatments are graph-derived choices under constructed constraints rather
than guideline-ranked clinical recommendations. The current task does not model
dosage, disease severity, patient preferences, local practice, guideline
changes, or care sequencing.

The benchmark also covers only one narrow task: choosing a treatment from a
fixed list after patient context changes. It does not cover history taking,
diagnosis, monitoring, escalation, or shared decision making. The vignettes and
rationales are synthetic benchmark materials, not clinical notes.

Results may also change with the candidate generator, prompt format, answer
parser, model version, and decoding settings. Future versions should add
broader clinical review, more disease areas, guideline-date checks, and
calibration analysis. Except for the controlled fine-tuning results where
standard errors are reported, the reported numbers are point estimates rather
than confidence intervals. We do not study reinforcement learning here; the
training experiments use supervised fine-tuning and replay.

%% file: sections/ethical_considerations.tex
\textsc{ClinPivot} is for research on model behavior, not for medical advice or
clinical deployment. The dataset uses public biomedical resources and synthetic
patient vignettes, not real patient records, which reduces privacy risk. The
main risk is misuse: benchmark performance could be mistaken for evidence that
a model is safe for patient care.

We therefore release the benchmark with documentation, source provenance,
validation scripts, and warnings that the examples are synthetic and incomplete.
Any model trained or evaluated with this benchmark would still need expert
clinical review before patient-facing or clinician-facing use.

Release documentation states the licenses and terms for source assets. We do
not redistribute raw PrimeKG files unless their terms permit it; released rows
are derived artifacts with source citations and metadata. Subject to source
terms, the derived benchmark is released under CC BY 4.0. We also disclose that
AI assistants were used for coding and editorial support. LLMs are used only as
a reject-only consistency screen and cannot assign gold labels.

%% file: sections/appendix.tex
\section{Benchmark Generation Details}
\label{app:generation}

\paragraph{Benchmark object.}
Each \textsc{ClinPivot} item is a treatment-selection problem built around a
disease $d$, an initially plausible treatment $a_0$, and a patient-specific
decision pivot $p$. The released row contains a natural-language multiple-choice
prompt \texttt{question}, a candidate set \texttt{choices}, the gold treatment
\texttt{answer} under graph-derived constraints, its zero-indexed position
\texttt{answer\_index}, a structured reference rationale \texttt{reasoning},
and metadata for audit and analysis.
At evaluation time, the model sees only the prompt and answer options; graph
node identifiers, pivot labels, banned-drug metadata, and evidence triples are
not shown to the model.

\paragraph{Construction workflow.}
The generator creates benchmark rows in five steps. First, it selects a disease
with at least one graph-supported indicated treatment and samples an initial
treatment $a_0$. Second, it searches for a feasible pivot $p$ tied to graph
relations or synthetic-but-labeled allergy constraints. Third, it renders the
pivot into the patient vignette, producing a new decision state $(d,p)$. Fourth,
it recomputes the gold treatment under graph-derived constraints for that
state. Finally, it builds a candidate set containing the post-pivot
graph-derived gold treatment, the pre-pivot treatment when it is useful as a
hard negative, and additional plausible alternatives. Rows are skipped when a
pivot cannot be grounded, when no valid post-pivot graph-derived gold can be
selected, or when the resulting item fails structural validation.

\paragraph{Gold-label semantics.}
For constraint pivots, the pivot makes the original treatment unsafe or
inappropriate, so the gold label is an indicated treatment for $d$ that is not
banned by the pivot. This label should be read as a graph-derived treatment
choice under the constructed constraints, not as a guideline-ranked clinical
recommendation. This directly tests whether the model moves away from the
pre-pivot disease--drug association. For off-label tradeoff pivots, the
indicated treatment remains the graph-derived gold while the off-label option
becomes a tempting distractor. This tests whether the model resists changing to
a less appropriate option when a related but insufficiently supported treatment
is introduced. Thus, \textsc{ClinPivot} includes both ``change when
constrained'' and ``do not change for the wrong reason'' cases.

\paragraph{Source graph and clinical text.}
The generator uses PrimeKG-style node metadata, disease--drug relations,
disease--phenotype relations, and clinical feature tables. Disease feature
fields provide textual clinical descriptions, symptoms, risk factors,
complications, causes, and management notes. Drug feature fields provide
descriptions, indications, mechanisms, pharmacodynamics, ATC codes, categories,
and related metadata. The graph relations identify candidate treatment edges and
decision pivots, while the feature fields supply natural-language context for
the vignette and metadata for plausible distractor selection.

\paragraph{Vignette construction.}
For a disease node, the generator renders a short synthetic clinical vignette
from a demographic header, disease feature text, linked phenotype concepts when
available, and the pivot text. The benchmark prompt is stored as
\texttt{question}, the gold drug as \texttt{answer}, the option list as
\texttt{choices}, and the zero-indexed gold option as \texttt{answer\_index}.

\paragraph{Pivot families.}
We instantiate five pivot families.
\begin{itemize}
    \item \textbf{Contraindication.} If the original drug is contraindicated
    for another disease, the vignette adds a history of that disease and marks
    the original drug as unsafe.
    \item \textbf{Comorbidity conflict.} The vignette adds a comorbid
    condition for which the original drug is contraindicated.
    \item \textbf{Drug--drug interaction.} The vignette adds a current
    medication that has a graph-backed interaction with the original drug. We
    treat this family as optional and include it only when the relation source
    supports a clinically constraining interpretation.
    \item \textbf{Allergy.} The vignette adds a deterministic synthetic allergy
    to the original drug or an inferred medication class. These rows are labeled
    as synthetic in metadata.
    \item \textbf{Off-label tradeoff.} The vignette introduces an off-label
    tempting alternative whose graph evidence supports use in a related
    condition but not as the preferred treatment for the current disease.
\end{itemize}

\paragraph{Candidate construction.}
Each item contains up to six answer choices. The choice set always includes the
gold drug. When a pivot bans the original drug, the banned drug is included as a
hard negative whenever it is present in the drug table. Additional choices are
drawn from other indicated drugs, contraindicated drugs, off-label drugs, and
pharmacologically nearby drugs. Nearness is approximated using drug metadata
such as ATC hierarchy, category, and group fields before falling back to broader
drug-description filters.

\paragraph{Reference rationale and metadata.}
Each row contains a \texttt{reasoning} field, but this field is not used as a
gold chain of thought. It is a structured reference rationale derived from the
same graph-backed note used to construct the item. The row metadata stores the
disease node, gold drug node, candidate drug nodes, pivot type, banned drugs,
evidence triples, and a structured JSON note.

\paragraph{Clinical consistency screen.}
For release artifacts, we add an LLM-assisted clinical consistency screen after
the complete item has been generated. The screen sees the vignette, choices,
graph-derived gold answer, pivot text, pivot effect, and evidence metadata. It
samples repeated judgments with temperature 0.9 and returns \texttt{keep},
\texttt{reject}, or \texttt{uncertain} plus short issue tags: an item is kept
when a majority of judgments consider the constructed example sound, rejected
when a majority do not, and marked uncertain when the judgments split evenly.
The screen is not allowed to relabel examples: rejected or uncertain rows are
removed from the release set, while kept rows preserve the original
graph-derived answer. We use \texttt{openai/gpt-oss-120b} for this screen and
also tested \texttt{Qwen/Qwen3.5-27B}, which produced similar aggregate
outcomes. We record the screen output in metadata and report the
keep/reject/uncertain counts. In the current release pass, the screen keeps
87\% of examples, rejects 9\%, and marks 4\% as uncertain; uncertain rows are
also excluded from the release set.

\paragraph{Splits and validation.}
Decision examples are split by disease node, so all pivoted rows for a disease
appear in the same split. This prevents near-duplicate disease contexts from
appearing in both training and test sets. The current release snapshot contains
20{,}000 examples, with 17{,}985 training examples and 2{,}015 test examples.
It spans 527 disease nodes, 1{,}320 gold treatment drugs, and 2{,}099 candidate
drugs. The train split contains 472 disease nodes and the test split contains
55 disease nodes, with no disease-node overlap. We run a structural
validator that checks that the gold answer appears in the options, that
\texttt{answer\_index} matches \texttt{answer}, that each row contains
graph evidence, that train/test disease leakage is absent, and that
pivot text makes banned drugs visible when a pivot bans an option. Validation
errors are treated as release blockers.

\begin{table}[t]
\centering
\small
\caption{\textsc{ClinPivot} release statistics.}
\label{tab:dataset_stats}
\begin{tabular}{lr}
\toprule
\textbf{Statistic} & \textbf{Count} \\
\midrule
Total examples & 20{,}000 \\
Train examples & 17{,}985 \\
Test examples & 2{,}015 \\
Disease nodes & 527 \\
Gold treatment drugs & 1{,}320 \\
Candidate drugs & 2{,}099 \\
\midrule
Contraindication pivots & 3{,}379 \\
Comorbidity-conflict pivots & 3{,}327 \\
Drug--drug interaction pivots & 5{,}201 \\
Allergy pivots & 5{,}653 \\
Off-label tradeoff pivots & 2{,}440 \\
\bottomrule
\end{tabular}
\end{table}

\paragraph{Scoring.}
Models are evaluated by feeding only the benchmark \texttt{question}. A response
is correct if the selected option matches the gold \texttt{answer\_index};
when an option index cannot be parsed, the evaluator falls back to normalized
drug-name matching against \texttt{answer}. Explanations may be generated for
qualitative analysis, but they are ignored for automatic accuracy unless they
change the final selected answer.

\section{Experimental Details}
\label{app:experimental-details}

\paragraph{Fine-tuning data.}
The controlled training experiments compare three supervision conditions. The
base condition evaluates the pretrained or instruction-tuned model without
additional task-specific clinical fine-tuning. The QA SFT condition fine-tunes
on PrimeKG-derived QA examples rendered from disease and drug feature fields.
The Decision SFT condition fine-tunes on \textsc{ClinPivot}-style decision
examples generated from the same source tables and graph relations. Thus, QA
and Decision SFT expose the model to overlapping disease--drug facts while
varying whether supervision is answer-oriented recall or action selection under
constraints.

\paragraph{Fine-tuning configuration.}
Unless otherwise specified, Qwen-3 experiments use full fine-tuning rather than
LoRA. We fine-tune for three epochs with learning rate $1\times10^{-6}$ and
batch size 8. Training jobs used either 8 Google Cloud TPU v6e accelerators or
4 NVIDIA A100 GPUs, depending on the run. Pretokenizing the training data took
approximately one hour. Fine-tuning runs took approximately 10--20 minutes once
the data were pretokenized. We did not perform hyperparameter search; all
training recipes use the fixed settings above, and we report the final
checkpoint after three epochs.

\paragraph{Retention setting.}
Clinical decision training can improve domain performance while eroding general
abilities that are important for deployable assistants, such as instruction
following, formatting, commonsense reasoning, and quantitative reasoning. We
therefore evaluate mixed QA + Decision SFT and replay variants. Replay mixes a
set of general instruction and reasoning examples, equal in size to the
clinical fine-tuning data, sampled from
\texttt{allenai/dolma3\_mix-6T}~\citep{olmo2025olmo3} into the clinical
decision-training stream, rather than replaying examples from the retention
evaluation sets. We use this as a standard mitigation strategy for forgetting
in sequential or task-adaptive
training~\citep{rolnick2019experiencereplaycontinuallearning,wu2024continuallearninglargelanguage,cho2026forgetforgettingcontinuallearning}.
General retention is summarized with IFEval~\citep{ifeval} and a broader
battery covering ARC, TruthfulQA, WinoGrande, GSM8K, HellaSwag, and MMLU.

\begin{table*}[t]
\centering
\small
\caption{Per-task general retention results underlying the aggregate retention
score in \Cref{tab:continual_learning}. Values are accuracies in percent.}
\label{tab:retention_full}
\begin{tabular}{lrrrrrrr}
\toprule
\textbf{Training Data} & \textbf{ARC} & \textbf{TruthfulQA} & \textbf{WinoGrande} & \textbf{GSM8K} & \textbf{HellaSwag} & \textbf{MMLU} & \textbf{Avg.} \\
\midrule
Base model & 70.19 & 50.46 & 69.90 & 58.76 & 77.64 & 69.80 & 66.10 \\
QA SFT & 67.12 & 47.84 & 66.30 & 55.22 & 75.00 & 69.12 & 63.42 \\
Decision SFT & 63.23 & 44.12 & 62.80 & 50.46 & 72.01 & 66.48 & 59.83 \\
QA + Decision SFT & 62.44 & 43.26 & 61.50 & 49.07 & 71.30 & 66.36 & 58.96 \\
Decision SFT + Replay & 68.91 & 49.58 & 68.40 & 56.29 & 76.05 & 70.22 & 64.87 \\
QA + Decision SFT + Replay & 68.12 & 48.82 & 67.71 & 55.40 & 75.52 & 70.96 & 64.41 \\
\bottomrule
\end{tabular}
\end{table*}

\begin{figure}[t]
\centering
\includegraphics[width=\linewidth]{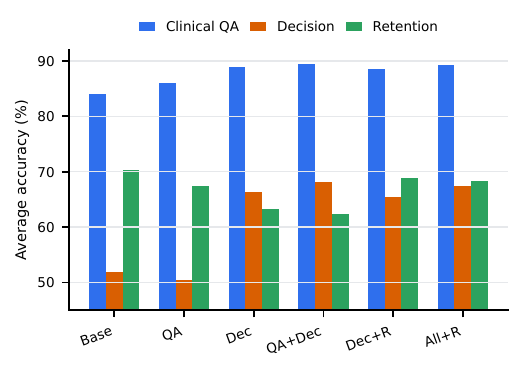}
\caption{Replay recovers much of the general retention lost during clinical
decision training. Each bar visualizes \Cref{tab:continual_learning} by
averaging the two metrics in the corresponding evaluation family.}
\label{fig:continual_learning}
\end{figure}

\paragraph{Closed-model evaluation.}
Closed-model evaluations were run through a hosted API evaluation service
available to the authors. We report public model names and shared decoding and
scoring settings, but exact backend checkpoint revisions for closed models are
not fully controlled by the authors.

\paragraph{Scoring protocol.}
For multiple-choice and candidate-selection settings, a response is scored
correct if the selected option matches the graph-derived gold label after
normalizing option letters and option text. For pivoted examples, decision
accuracy is computed against the recomputed gold treatment under graph-derived
constraints. Pivot sensitivity is computed from whether the model avoids the
pre-pivot treatment when that treatment is explicitly banned by the pivot
metadata. Evaluation uses deterministic decoding
where supported, with temperature 0.0, top-$p$ 1.0, and a maximum of 256 new
tokens. For \Cref{tab:fine_tuning_results}, we report means and standard errors
across five random seeds; other reported experimental results are point
estimates unless otherwise specified.